# Transformer Models for Text Summarization: A Comparative Study of BART, BERT, and RoBERTa

Daisy Aptovska and Vinayak Elangovan
Computer Science program, Penn State University Abington, PA, USA

## *ABSTRACT*

*Text summarization refers to the task of condensing a document into a shorter version while preserving its–key information. Automatic text summarization (ATS), driven by advancements in natural language processing (NLP), has developed rapidly in recent years. ATS methods are commonly categorized by input type (such as single-document or multi-document summarization) and by output type (extractive, abstractive, and hybrid). This article presents a focused review of modern summarization techniques with an emphasis on transformer-based models and large language models (LLMs), specifically BERT, RoBERTa and BART. It examines their architectures, pretraining strategies, and their suitability for extractive and abstractive summarization tasks. The paper also discusses key challenges, including computational requirements, data limitations, and issues such as factual inconsistency in generated summaries, and highlights the strengths and limitations of encoder-only and encoder–decoder models.*

## 1. INTRODUCTION

The rapid growth of digital text across domains such as scientific research, journalism, legal documentation, and social media has created a strong need for efficient methods to process and understand large volumes of information. Automatic Text Summarization (ATS) addresses this challenge by generating concise summaries that preserve the most important content, supporting applications in information retrieval, knowledge management, and decision-making. Early approaches to ATS relied on statistical and rule-based techniques, which were computationally efficient but often lacked contextual understanding and linguistic coherence. With the advancement of deep learning, neural network-based methods significantly improved summarization quality. More recently, transformer-based models have become the dominant approach in natural language processing, enabling substantial progress in both extractive and abstractive summarization tasks.

Pre-trained language models such as BERT (Bidirectional Encoder Representations from Transformers), RoBERTa (A Robustly Optimized BERT Pretraining Approach), and BART (Bidirectional and Auto-Regressive Transformers) have demonstrated strong performance across a wide range of NLP applications. Despite sharing a common transformer architecture, these models differ in their design and pretraining objectives. BERT and RoBERTa are encoder-only models primarily suited for text understanding tasks, while BART employs an encoder-decoder architecture that is more effective for text generation tasks such as abstractive summarization. These differences make them suitable for different types of summarization approaches and application scenarios.

Although several surveys have reviewed ATS methods and transformer-based models, many focus broadly on summarization techniques or large language models without providing a detailed comparison of specific transformer architectures for summarization. In particular, there is limited discussion on how architectural differences between encoder-only and encoder–decoder models influence their performance and suitability for extractive and abstractive tasks.

This paper presents a focused review of transformer-based summarization methods, with a comparative analysis of BERT, RoBERTa, and BART. It examines their architectures, pretraining strategies, and applications in ATS, and discusses their strengths, limitations, and practical considerations. The goal is to provide a clearer understanding of how these models differ and to support the selection of appropriate approaches for different summarization tasks and resource settings.

This paper is organized as follows. Section-2 presents a review of related work and foundational concepts in Automatic Text Summarization (ATS). Section-3 describes the methodologies employed in the study, including dataset selection, preprocessing procedures, and model training. Section-4 examines the challenges encountered during the implementation process, along with relevant challenges reported in prior research. Section-5 details the experimental results and evaluates model performance using the Recall-Oriented Understudy for Gisting Evaluation (ROUGE) metric. Section-6 summarizes the key findings and outlines potential directions for future work.

## 2. RELATED WORKS

A substantial body of research has explored Automatic Text Summarization (ATS) from both methodological and application-oriented perspectives. Several comprehensive surveys have reviewed the evolution of ATS techniques, covering statistical, machine learning, and deep learning approaches [1], [5], [12]. These studies provide broad overviews of the field, highlighting the transition from extractive to abstractive methods and the growing role of neural architectures.

In the area of extractive summarization, prior work has focused on improving sentence selection using feature-based and graph-based approaches. For example, comparative studies have analyzed the effectiveness of features such as term frequency, sentence position, and similarity measures in multi-document summarization [2], while more recent approaches utilize graph representations and attention mechanisms to enhance relevance and coherence [4]. Although these methods improve content selection, they are inherently limited to extracting existing sentences and often fail to produce fluent summaries.

Abstractive summarization has been extensively studied as a means to overcome these limitations. Early neural approaches employed sequence-to-sequence models based on recurrent and convolutional architectures to generate summaries with improved fluency and semantic representation [7]. More recent work has explored deep learning methods for low-resource and domain-specific summarization tasks, emphasizing the importance of data quality and model generalization [3]. Despite these advances, challenges such as data dependency, scalability, and factual inconsistency remain significant concerns.

The introduction of transformer-based models has significantly advanced the state of the art in ATS. BERT introduced bidirectional contextual representations that improved performance in text understanding and extractive summarization tasks [8]. RoBERTa further enhanced these capabilities through optimized pretraining strategies and large-scale data utilization [9]. BART extended transformer architectures by incorporating an encoder–decoder framework, enabling strong performance in generative tasks such as abstractive summarization [10]. Recent studies have also explored hybrid and integrated transformer-based approaches for improved summarization performance [12].

While existing surveys provide valuable insights into ATS techniques and model developments, most adopt a broad perspective and do not focus on detailed comparisons between specific transformer architectures. In particular, there is limited analysis of how differences in model design—such as encoder-only versus encoder–decoder structures—and pretraining objectives impact their effectiveness in extractive and abstractive summarization. This gap motivates the need for a focused comparative review of models such as BERT, RoBERTa, and BART.

# 3. FOUNDATIONAL CONCEPTS

Automatic Text Summarization (ATS) is the process of generating a concise version of a document while preserving its key information. ATS methods are generally categorized into extractive and abstractive approaches [1], [5].

Extractive summarization selects important sentences or phrases directly from the source text based on statistical or learned features. Common techniques include term frequency-based methods and graph-based approaches such as TextRank and LexRank, which model relationships between sentences to determine their importance [2], [4]. While extractive methods are computationally efficient and preserve factual accuracy, they often lack coherence and readability.

Abstractive summarization, in contrast, generates new sentences that paraphrase the original content. This approach typically uses sequence-to-sequence models and deep learning architectures to capture semantic meaning and produce more natural summaries [7], [3]. However, abstractive methods require large training datasets and may introduce errors or factual inconsistencies.

Transformer-based models have become the dominant approach in modern ATS due to their ability to capture long-range dependencies using self-attention mechanisms [6]. Models such as BERT are primarily used for text understanding and extractive tasks, while encoder–decoder models such as BART are better suited for generative tasks like abstractive summarization [8], [10]. RoBERTa improves upon BERT by optimizing training strategies and using larger datasets, resulting in improved performance across various NLP tasks [9].

The performance of summarization models is typically evaluated using automated metrics such as ROUGE and BLEU, which measure the similarity between generated summaries and reference summaries [11]. While these metrics provide quantitative evaluation, they may not fully capture semantic quality and human readability.

# 4. METHODOLOGIES

This section outlines the experimental framework used to support the comparative analysis of transformer-based models for automatic text summarization. The goal is not to propose new models but to provide practical insights into the performance and behavior of BERT, RoBERTa, and BART under a consistent evaluation setting.

## 4.1 Dataset Selection

The study uses the widely adopted CNN/Daily Mail dataset (obtained from Kaggle in 2025), consisting of news articles paired with human-written summaries. These reference summaries are used to evaluate model-generated outputs using ROGUE and BLEU metrics. This dataset presents a realistic challenge for both extractive and abstractive summarization models due to its diverse content and variations in article length. Data was split into training, validation, and test subsets following established standards. Due to hardware limitations, a reduced subset of the dataset was used to enable training and evaluation on a CPU-based system. Specifically, 15,000 samples samples from the training set, 2,000 from the validation set, and 100 from the test set were selected. While this limits large-scale generalization, it allows for a controlled comparison of model behavior under constrained computational resources.

## 4.2 Preprocessing and Tokenization

Preprocessing was customized to meet the specific requirements of each model. This process involved normalizing the text by converting all characters to lowercase and removing extraneous whitespace. Transformer-based models generally require minimal preprocessing, as they are designed to capture contextual relationships directly from raw text. Tokenization was performed using model-specific tokenizers available in the Hugging Face Transformers library, with WordPiece applied for BERT, and Byte-Pair Encoding (BPE) used for RoBERTa, and BART. To ensure consistency and efficient processing, input sequences were truncated or padded to fixed length of 512 tokens for BERT and RoBERTa, and up to 1024 tokens for BART. Special tokens required for each model architecture were incorporated as needed.

## 4.3 Model Architectures and Implementation

### 4.3.1 BERT

The BERT-based summarization approach employed a pretrained BERT Base model adapted for extractive summarization. A sentence-level classification layer was added to identify and select important sentences from the input text. The model was fine-tuned using cross-entropy loss on the labeled dataset.

### 4.3.2 RoBERTa

RoBERTa's architecture is similar to BERT but incorporates improved pretraining strategies. RoBERTa was integrated into the BERT framework to leverage its contextual embeddings for extractive summarization. Fine-tuning was performed using comparable training procedures and loss functions to maintain consistency across models.

### 4.3.3 BART

BART is a sequence-to-sequence model pretrained as a denoising autoencoder. Its encoder-decoder architecture supports abstractive summarization by generating new text based on the input document. In this study, a pretrained BART model optimized for summarization was used to generate summaries, and its outputs were evaluated against reference summaries using standard metrics.

## 4.4 TRAINING PIPELINES AND COMPATIBILITY

All models were implemented using Python 3.12 with the Hugging Face Transformers library and PyTorch as the deep learning framework. Training and inference were conducted on a CPU-based system without GPU acceleration, reflecting a resource constrained experimental setup.

For BERT and RoBERTa, training involved fine-tuning pretrained models on the selected dataset subset using standard training configurations. Due to computational limitations, reduced batch sizes and a limited number of training epochs were used.

BART was evaluated using a pretrained model without additional fine-tuning due to computational constraints. Therefore, the comparison reflects differences between fine-tuned encoder-only models and a pretrained encoder–decoder model designed for summarization, rather than a fully controlled training setup. While this introduces differences in training conditions, it reflects practical usage scenarios and is considered when interpreting results.

### 4.5 Evaluation Metrics

Model performance was evaluated using widely used automatic metrics, including ROUGE and BLEU. ROUGE measures n-gram overlap between generated summaries and reference summaries, while BLEU evaluates precision-based similarity. These metrics provide a quantitative basis for comparing summarization performance across models. Although widely adopted, these metrics have limitations in capturing semantic coherence and factual accuracy, and therefore results are interpreted with these constraints in mind.

## 5. CHALLENGES AND LIMITATIONS

During the implementation and comparative analysis of transformer-based models for automatic text summarization, several technical and methodological challenges were encountered. These challenges are closely related to computational constraints, model architecture differences, dataset limitations, and known limitations of generative language models.

A primary challenge was the limited availability of computational resources. Transformer-based models such as BERT, RoBERTa, and BART are computationally intensive due to their large number of parameters and reliance on self-attention mechanisms. In this study, all experiments were conducted on a CPU-based system without GPU acceleration, which significantly constrained training efficiency. As a result, smaller batch sizes, reduced sequence lengths, and a limited number of training epochs were used to ensure feasibility. These adjustments are consistent with findings in prior research, which show that reducing batch size and sequence length can help mitigate memory constraints in resource-limited environments [13].

Another key challenge relates to the architectural differences between encoder-only and encoder–decoder models, which directly impacted implementation consistency. BERT and RoBERTa are inherently designed as encoder-only models and are primarily suited for extractive summarization tasks, whereas BART follows an encoder–decoder architecture specifically designed for sequence generation and abstractive summarization. This difference made it difficult to maintain identical training and evaluation pipelines across models. In particular, attempts to adapt BERT for generative decoding tasks were not feasible due to its architecture limitations, reinforcing its suitability for extractive summarization rather than generation-based approaches.

A well-known limitation in transformer-based language models is the issue of factual inconsistency or hallucination, particularly in abstractive summarization. Although BART is pretrained as a denoising autoencoder and performs well in generating fluent summaries, it may still produce content that is not fully aligned with the source text. This issue is less prominent in extractive approaches (BERT and RoBERTa), as they rely on selecting existing sentences rather than generating new content. This trade-off between fluency and factual reliability remains a key challenge in comparing extractive and abstractive summarization methods.

Another significant challenge is dataset dependency and data limitations. Although the CNN/DailyMail dataset is widely used for summarization research, constructing high-quality summarization datasets remains difficult due to the need for human-written reference summaries. Additionally, the use of a reduced subset of the CNN/DailyMail dataset due to computational constraints limits the generalizability of the results, and performance may differ when evaluated on the full dataset or larger-scale benchmarks. Prior work has also highlighted the scarcity and high cost of creating large-scale, high-quality summarization datasets [1].

Finally, during experimentation, we encountered implementation challenges in adapting pretrained models for consistent comparison. While BERT and RoBERTa were fine-tuned for extractive summarization using sentence-level classification, BART was used in its pretrained form without additional fine-tuning due to resource limitations. This introduced a difference in training conditions, which was carefully considered during evaluation and interpretation of results. These constraints highlight a broader challenge in comparative NLP studies: ensuring methodological parity across fundamentally different model architectures while working within limited computational resources.

To further examine model behavior beyond quantitative metrics, a qualitative analysis was conducted on a subset of summaries generated by BART. While the model consistently produced fluent and coherent summaries, occasional instances of factual inconsistency were observed during manual inspection. Some generated summaries included minor deviations from the source text, such as slight alterations in numerical values, omission of key contextual qualifiers, or the introduction of loosely inferred details. These inconsistencies were not pervasive but highlight a known limitation of abstractive summarization models, where the generation process may prioritize fluency and compression over strict factual alignment with the source document. In contrast, extractive models such as BERT and RoBERTa preserved factual accuracy by selecting sentences directly from the original text. However, this often resulted in summaries that were less concise and less coherent compared to those generated by BART.

## 6. RESULTS AND DISCUSSION

After 16 epochs of training, the BERT-based extractive summarization model achieved a final training loss of 2.11422, while the RoBERTa model achieved a lower training loss of 1.47048, indicating better convergence under the same training configuration. In terms of computational cost, BERT required approximately 4 hours to complete 16 epochs of training, whereas RoBERTa required nearly 5 hours, reflecting the additional computational overhead associated with its larger pretraining corpus and optimized architecture.

In contrast, the BART model was evaluated using a pretrained checkpoint (*BART-large-CNN*) without additional fine-tuning due to computational constraints. As a result, its inference process was significantly faster and more efficient. For all three models, inference time remained under 10 minutes for generating 100 summaries from the test subset, demonstrating that summarization at inference time is computationally manageable even under CPU-based execution.

### 6.1 BERT vs. RoBERTa

Across both ROUGE and BLEU evaluation metrics, RoBERTa consistently outperformed BERT under identical fine-tuning conditions. Although both models were trained using the same extractive summarization framework, RoBERTa benefited from improved pretraining strategies, including larger training corpora and optimized training procedures. These enhancements enabled RoBERTa to generate more contextually relevant sentence representations, leading to improved sentence selection for summarization tasks. This performance gap highlights the impact of pretraining scale and optimization even when downstream training setups are held constant.

## 6.2 BERT vs. BART

A clear performance difference was observed between BERT and BART. While BERT is an encoder-only model designed primarily for understanding tasks, BART employs an encoder–decoder architecture specifically designed for sequence generation. This architectural advantage enables BART to perform abstractive summarization more effectively by generating coherent and fluent summaries rather than selecting existing sentences. Additionally, BART is pretrained as a denoising autoencoder on large-scale corpora, making it inherently more suitable for summarization tasks. As reflected in **Table 1**, BART significantly outperformed BERT across all ROUGE and BLEU metrics.

## 6.3 RoBERTa vs. BART

When comparing RoBERTa and BART, the results indicate that BART achieved substantially higher performance even without task-specific fine-tuning, whereas RoBERTa required 16 epochs of supervised fine-tuning for extractive summarization. Specifically, BART achieved a ROUGE-1 score of approximately 0.4117, compared to 0.1820 for RoBERTa. Similar trends were observed across ROUGE-2, ROUGE-L, ROUGE-LSum, and BLEU metrics.

This suggests that pretrained encoder–decoder architectures such as BART have a strong inherent advantage in summarization tasks due to their generative capabilities and large-scale pretraining. However, it is important to note that this comparison reflects different training conditions, which should be considered when interpreting results.

Table 1. Evaluation Results of BERT, RoBERTa, and BART

| **Model** | **ROUGE-1** | **ROUGE-2** | **ROUGE-L** | **ROUGE-LSum** | **BLEU** |
|---|---|---|---|---|---|
| BERT | 0.14854 | 0.02340 | 0.11024 | 0.13769 | 0.01633 |
| RoBERTa | 0.18200 | 0.03144 | 0.14114 | 0.16474 | 0.02523 |
| BART | 0.41173 | 0.19206 | 0.28982 | 0.35358 | 0.16256 |

## 6.4 Interpretation of Results

The results in Table 1 highlight that model performance is strongly influenced by architectural differences and pretraining strategies. RoBERTa consistently improves over BERT across all evaluation metrics, indicating that enhanced pretraining methods lead to better contextual representations for extractive summarization. A more significant performance gap is observed between encoder-only models (BERT and RoBERTa) and the encoder–decoder model (BART). This difference is primarily due to the generative capability of BART, which allows it to produce abstractive summaries rather than selecting existing sentences. This leads to better semantic compression and higher overlap with reference summaries, as reflected in the evaluation scores. However, this improvement should be interpreted with caution since BART was evaluated in a pretrained (non-fine-tuned) setting, whereas BERT and RoBERTa were fine-tuned on the dataset subset. Despite this, the results still clearly demonstrate the advantage of encoder–decoder architectures for abstractive summarization tasks.

Overall, the findings suggest that RoBERTa improves upon BERT due to stronger pretraining strategies, while BART achieves the highest performance due to its encoder–decoder generative architecture. Extractive models (BERT and RoBERTa) are more stable and fact-preserving, whereas abstractive models (BART) produce more fluent and semantically rich summaries but depend heavily on pretraining and generation behavior.

## 7. CONCLUSION

This study analyzed the performance of three transformer-based models—BERT, RoBERTa, and BART—for the task of Automatic Text Summarization (ATS), under a controlled experimental setting. The models were evaluated using standard ROUGE and BLEU metrics on a subset of the CNN/DailyMail dataset, with BERT and RoBERTa fine-tuned for extractive summarization and BART evaluated using a pretrained abstractive summarization model.

The results demonstrate that model architecture and pretraining strategy play a critical role in summarization performance. Among the three models evaluated, BART consistently achieved the strongest performance, significantly outperforming both BERT and RoBERTa across all evaluation metrics. This can be attributed to its encoder–decoder architecture and large-scale denoising pretraining, which make it inherently well-suited for abstractive text generation tasks. In contrast, RoBERTa outperformed BERT under the same fine-tuning conditions, highlighting the benefits of improved pretraining strategies and larger training corpora even within encoder-only architectures. BERT, while effective for extractive summarization, showed comparatively lower performance, reflecting its more limited pretraining scale and design constraints for generative tasks.

Overall, the findings from this study reinforce several key observations. First, RoBERTa improves upon BERT due to stronger pretraining strategies that enhance contextual representation learning. Second, encoder–decoder architectures such as BART provide a clear advantage for abstractive summarization by enabling fluent and semantically rich text generation rather than sentence selection. Third, extractive models such as BERT and RoBERTa tend to be more fact-preserving and stable, while abstractive models like BART achieve higher semantic quality but are more dependent on generative pretraining behavior. However, it is also important to note that the comparison is influenced by differences in training configurations, particularly the use of fine-tuned encoder-only models versus a pretrained generative model. This should be considered when interpreting the magnitude of performance differences.

Future work can extend this study by fine-tuning BART on domain-specific datasets such as medical, legal, or educational corpora to improve specialization and robustness. Additionally, further research could explore comparisons with more recent large language models such as T5, LLaMA, and GPT-based architectures, which offer significantly larger parameter scales and may further advance the state of the art in automatic text summarization.

# References

1. Sharma, G., & Sharma, D. (2023). Automatic Text Summarization Methods: A Comprehensive Review. *SN Computer Science*, 4(1), 33.
2. Mutlu, B., Sezer, E. A., & Akcayol, M. A. (2019). Multi-document extractive text summarization: A comparative assessment on features. *Knowledge-Based Systems*, 183, 104848.
3. Shafiq, N., Hamid, I., Asif, M., Nawaz, Q., Aljuaid, H., & Ali, H. (2023). Abstractive text summarization of low-resourced languages using deep learning. *PeerJ Computer Science*, 9, e1176.
4. Lin, Y. C., & Ma, J. (2023). Automatic Text Extractive Summarization Based on Text Graph Representation and Attention Matrix. *Lecture Notes in Computer Science*, vol. 14089. Springer.

5. Widyassari, A. P., et al. (2022). Review of automatic text summarization techniques & methods. *Journal of King Saud University – Computer and Information Sciences*, 34(4), 1029–1046.
6. Khurana, D., et al. (2023). Natural language processing: state of the art, current trends and challenges. *Multimedia Tools and Applications*, 82, 3713–3744.
7. Song, S., Huang, H., & Ruan, T. (2019). Abstractive text summarization using LSTM-CNN based deep learning. *Multimedia Tools and Applications*, 78(1), 857–875.
8. Devlin, J., et al. (2018). BERT: Pre-training of Deep Bidirectional Transformers for Language Understanding. *NAACL-HLT*, 4171–4186.
9. Liu, Y., et al. (2019). RoBERTa: A Robustly Optimized BERT Pretraining Approach. *arXiv:1907.11692*.
10. Lewis, M., et al. (2019). BART: Denoising Sequence-to-Sequence Pre-training for Natural Language Generation, Translation, and Comprehension. *ACL*, 7871–7880.
11. Menéndez, H. D., & Dakhama, A. (2025). Automatic Summarization Evaluation: Methods and Practices. *Lecture Notes in Computer Science*, vol. 15383. Springer.
12. Yu, P. S., Zhang, H., & Zhang, J. (2024). A Systematic Survey of Text Summarization: From Statistical Models to Large Language Models. *arXiv:2406.11289*.
13. Areshey, A. & Mathkour, H. (2024). *Exploring transformer models for sentiment classification: A comparison of BERT, RoBERTa, ALBERT, DistilBERT, and XLNet.* Expert Systems, 41(11), e13701.

## Appendix A: Implementation Details

The implementation pipeline involved standard data preparation and model training procedures using the Hugging Face Transformers framework. The dataset was formatted into a structured dataset object after removing non-essential fields. Model-specific tokenizers were applied to convert text into token representations, followed by fine-tuning using appropriate training configurations for BERT and RoBERTa, and inference using a pretrained BART model. Model outputs were generated on the test dataset and evaluated against reference summaries using ROUGE and BLEU metrics.

| Component | BERT | RoBERTa | BART |
|---|---|---|---|
| **Model Type** | Encoder-only | Encoder-only | Encoder–decoder |
| **Summarization Approach** | Extractive | Extractive | Abstractive |
| **Pretraining Objective** | Masked Language Modeling (MLM) + Next Sentence Prediction | MLM (optimized training, no NSP) | Denoising autoencoder (sequence-to-sequence) |
| **Tokenizer** | WordPiece | Byte Pair Encoding (BPE) | Byte Pair Encoding (BPE) |
| **Input Representation** | Tokenized sentences with classification layer | Similar to BERT with improved embeddings | Full document input for sequence generation |
| **Max Input Length** | 512 tokens | 512 tokens | Up to 1024 tokens |
| **Training Strategy** | Fine-tuned on dataset subset | Fine-tuned on dataset subset | Pretrained model used (no fine-tuning) |
| **Training Configuration** | Limited epochs, small batch size (CPU-based) | Similar setup with adjusted epochs | Inference-based evaluation |
| **Summary Generation** | Sentence selection (classification scores) | Sentence selection (classification scores) | Text generation using decoder |

| Component | BERT | RoBERTa | BART |
|---|---|---|---|
| **Evaluation Metrics** | ROUGE, BLEU | ROUGE, BLEU | ROUGE, BLEU |
| **Computational Setup** | CPU-based environment | CPU-based environment | CPU-based environment |
| **Strength in Context** | Strong contextual understanding for extraction | Improved robustness over BERT | Strong generation and fluency |
| **Limitation in Context** | Limited to extractive summaries | Same limitations as BERT | Higher computational cost, potential hallucination |

**Appendix B: Specifications of CPU Used**

Apple Mac Studio with M4 Max (14-core CPU), 36 GB RAM, and 512 GB SSD storage.